\pdfoutput=1

\documentclass[11pt]{article}
\usepackage{amsmath}
\usepackage{mathtools}

\usepackage[]{acl} 

\usepackage{times}
\usepackage{latexsym}

\usepackage[T1]{fontenc}

\usepackage[utf8]{inputenc}

\usepackage{microtype}
\usepackage{multirow}
\usepackage{longtable}
\usepackage{stfloats}

\title{Latent Prompt Tuning for Text Summarization}

\author{Yubo Zhang$^{1}$\thanks{~~Work done during Yubo Zhang's internship at Microsoft Research Asia.}, ~~Xingxing Zhang$^{2}$, ~~Xun Wang$^{2}$, ~~Si-qing Chen$^{2}$ \and Furu Wei$^{2}$ \\
$^1$Department of Computing, The Hong Kong Polytechnic University \\
$^2$Microsoft \\
$^{1}$\texttt{yubo.zhang@connect.polyu.hk} \\
$^{2}$\texttt{\{xizhang,xunwang,sqchen,fuwei\}@microsoft.com} \\
}

\begin{document}
\maketitle
\begin{abstract}

Prompts with different control signals (e.g., length, keywords, etc.) can be used to control text summarization. When control signals are available, they can control the properties of generated summaries and potentially improve summarization quality (since more information are given). Unfortunately, control signals are not already available during inference time. In this paper, we propose {\sc Lotus} (shorthand for {\bf L}atent Pr{\bf O}mpt {\bf Tu}ning for {\bf S}ummarization), which is a \emph{single} model that can be applied in both ``controlled'' and ``uncontrolled'' (without control signals) modes. During training, {\sc Lotus} learns latent prompt representations from prompts with gold control signals using a contrastive learning objective. Experiments show {\sc Lotus} in uncontrolled mode consistently improves upon strong (uncontrollable) summarization models across four different summarization datasets. We also demonstrate generated summaries can be controlled using prompts with user specified control tokens.

\end{abstract}

\section{Introduction}
\label{sec:intro}

Abstractive summarization models aim to automatically paraphrase a longer document into a shorter version (i.e., a summary), and the generated summary is likely to contain new phrases and sentences that may not appear in the source article. The task of text summarization can greatly help people to seek information you need in a faster manner. Sometimes, a general purpose summary for a document is not sufficient, since users may prefer summaries containing certain keywords or summaries with length constraints. Controllable text summarization \cite{AngelaFan2017ControllableAS} is a sub-field in text summarization, which  focuses on enabling summarization models to generate summaries controlled by user-specific requirements. Controllable summarization seems to be a good supplement for general purpose summarization.

Typical controllable models \cite{AngelaFan2017ControllableAS,he2020ctrlsum} prepend control tokens (e.g., length of the target summary and keywords) to the beginning of its input document during training time. These control signals (tokens) can be obtained from a regular summarization dataset (without explicit control signal annotations). During test time, any valid control signal can be used as input. Generated text can be manipulated with different input control signals. These control signals essentially define the property of the desired text. During training time they can be also viewed as \emph{hints} for the output summary. For example, during training, the control signal \emph{length 80} informs the model to predict EOS (end of sentence) token at around the \emph{80}th word, which intuitively makes the generation task easier. Therefore, it is possible to use a model with control signals (more information regarding the gold text) to help a model without them. On the other hand, controllable model are trained with control signals, but it is still possible during test time no control signals are available. For example, when users do not know which query to use, but they are still interested to have a summary for a document.

Clearly, controlled and uncontrolled summarization models have their own advantages. So \emph{can we have a model that can do them all?} \emph{Can controllable and uncontrollable models help each other?} 

In this paper, we propose \textsc{Lotus} (shorthand for \textbf{L}atent Pr\textbf{O}mpt \textbf{Tu}ning for \textbf{S}ummarization), which is a \textit{single} model that can be applied in both ``controlled'' and ``uncontrolled'' modes. We achieve controllable summarization through  prompts. During training time, the uncontrolled model learns a \emph{latent prompt} from the controlled model using a contrastive learning objective. During test time, we can use different prompts to do controllable and uncontrollable generation. We conducted both experiments in both uncontrolled and controlled modes. Experiments on CNN/DailyMail, XSum, Gigaword, and New York Times show {\sc Lotus} in uncontrolled mode consistently improves upon strong (uncontrollable)  models. We also demonstrate that in controlled mode, generated summaries from {\sc Lotus} can be controlled using prompts with user-specified control signals.

\section{Related Work}

Large pre-trained Seq2Seq models can be used to generate high-quality summaries \cite{radford2019language,MichaelLewis2019BARTDS,zhang2020pegasus,raffel2020exploring}. But most of them are focusing on summarization with general purpose, and do not allow models to generate summaries controlled by user-specific requirements. As a sub-field in text summarization, controllable summarization \cite{AngelaFan2017ControllableAS} aims to enable models to generate summaries controlled by user-specific requirements. For example, some previous length control methods \cite{rush2015neural,kikuchi2016controlling,bian2019controllable}
can allow users to set their preferred summary length. Their controllable methods are focusing on the model decoding phase. For example, \citet{rush2015neural} stops the model decoding at a specified length by informing the model to generate an EOS token (i.e., via assigning $-\infty$ scores to other vocabulary tokens).

In addition to length control, \citet{AngelaFan2017ControllableAS} later proposed prepending control tokens (e.g., length of the target summary and entities) to the beginning of its input document can hint Seq2Seq model to generate summaries based on these control tokens. These control signals (tokens) can be obtained from the ground-truth summaries or specified by users. \citet{he2020ctrlsum} then propose CTRLSum that considers keywords as a kind of control signal, and keywords extraction is based on reference summaries during CTRLSum training. During CTRLSum inference, its keywords extraction is based on an additional BERT-based keywords extraction model that can extract keywords when the reference summaries are unavailable. \citet{xu-lapata-2022-document} models query-focused summarization (QFS) and general purpose summarization using a single model, which is similar to our method in the sense that our model can do controllable and uncontrollable summarization in single model. However, QFS is different from controllable summarization and our joint modeling, which are based on prompting and contrastive learning, does not introduce any additional model parameters.

Some approaches focused on modifying the training objectives to be conditioned on the control signals. \citet{earle2021learning} formulate the controllable generation as a reinforcement learning task by adding an additional reward if model's generation (i.e., action) can meet control signals. Recently, there are other methods such as \citet{carlsson2022fine} and \citet{liu2022length} that extend the model's pre-training objectives to incorporate control ability. For example, \citet{liu2022length} uses a length-aware attention mechanism to select samples with similar lengths based on the length control signal, which ability can be transferred to the downstream tasks. 

\begin{figure*}[!ht]
    \centering
    \includegraphics[width=\linewidth]{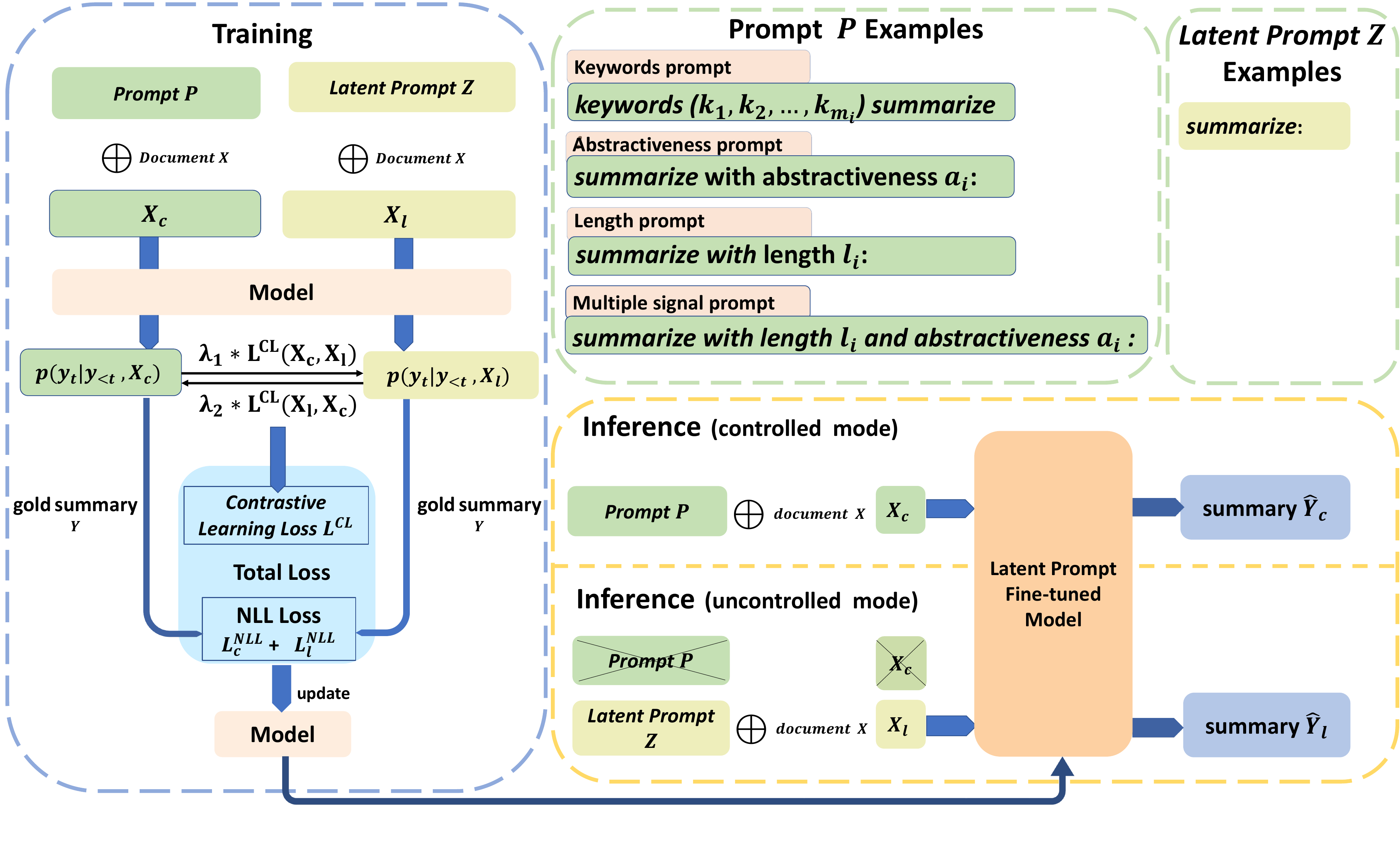}
    \caption{Overview of \textsc{Lotus}. Left is the training phase of \textsc{Lotus}. Examples of ``Prompt'' and ``Latent Prompt'' are shown in the upper right. Bottom right illustrates \textsc{Lotus} inference for both controlled and uncontrolled modes.}
    \label{fig:arch}
\end{figure*}

\section{Neural Abstractive Summarizer}
\label{sec:neusum}

Abstractive summarization aims to generate a short summary given a document, which can be viewed as a typical sequence-to-sequence learning problem (the input document and the output summary are two sequences of tokens). Given a document $X=\left(x_1, x_2, \ldots, x_{|X|}\right)$ and its gold summary $Y=\left(y_1, y_2, \ldots, y_{|Y|}\right)$, the training objective is to estimate the following conditional probability using a Transformer encoder-decoder model \cite{Vaswani2017NIPS}:
\begin{equation}
    \label{eq:model_prob}
    p(Y \mid X ; \theta)=\prod_{t=1}^{|Y|} p\left(y_t \mid y_{<t}, X ; \theta\right)
\end{equation}
where $\theta$ is the model parameter and $y_{<t}$ denotes all tokens before position $t$ (i.e., $\left.\left(y_1, y_2, \ldots, y_{t-1}\right)\right)$.

The Seq2Seq Transformer model can be trained by minimizing the negative log-likelihood (NLL) of gold document-summary pairs:
\begin{equation}
    \label{eq:nll_loss}
    \mathcal{L}^{\text{NLL}}(\theta)=-\frac{1}{|Y|} \log p(Y \mid X ; \theta)
\end{equation}
where $|Y|$ is the number of tokens in summary $Y$.

\section{Latent Prompt Tuning}
\label{sec:lotus_model}

In this section, we introduce our model \mbox{\sc Lotus}, which is a model can be used in both controllable and uncontrollable modes. Control signals (e.g., length of generated summaries, named entities that appears in summaries) are usually available during model training phase and are likely to help uncontrollable generation (since more information for summaries are given). We first introduce controllable generation and then introduce how controllable model can be used to help uncontrollable generation with a contrastive learning objective.

\subsection{Controllable Text Summarization}
\label{sec:ctrlsum}

As demonstrated by GPT-3 \cite{brown2020language} and T5 \cite{raffel2020exploring}, controllable text generation can be achieved through prompting. For example, to control lengths of the generation summaries, we can prepend the prompt ``{\tt summarize with length 30:}'' to the input document $X$. After training, we can generate a summary $Y$ with a length of around 30 tokens. We can still use the Seq2Seq Transformer model described in Section \ref{sec:neusum} to implement controllable text summarization, but the training examples need to be transformed as follows. Let ($X_i$, $Y_i$) denote a document-summary pair in a summarization dataset $\mathcal{D}$. After transformation, the example becomes ($P_i \bigoplus X_i $, $Y_i$) and the resulting dataset is denoted as $\mathcal{D}^{\text{P}}$, where $\bigoplus$ indicates concatenation and $P_i$ is a text prompt for document $X_i$. We then train a Transformer model on $\mathcal{D}^{\text{P}}$ using the NLL objective (Equation \ref{eq:nll_loss}) and the  model can do \mbox{controllable} generation. In the following, we describe how different prompt $P_i$ is constructed to achieve control of summary \mbox{generation}.

\paragraph{Length} Given a document, usually the length of the generated summary from a summarization model is fixed, which may be different from what users expected. We therefore propose to prepend the prompt ``{\tt summarize with length $l_i$}: '' to a document $X_i$ in order to control its summary length, where $l_i$ is the \emph{gold} number of tokens in the summary $Y_i$ during training, while during inference $l_i$ is given as a control signal. Summaries with a specific length (e.g., 98) in the training dataset may be very sparse. Therefore, we use length bins of size 5 instead of the original lengths to group summaries of similar lengths. For example, $l_i = 5$ means the length of the summary is between 5 and 10 tokens.

\paragraph{Abstractiveness} Abstractiveness measures the $n$gram overlap between a summary and its original document. Specifically, the more $n$grams are copied from the document, the less abstractive a summary is. The prompt ``{\tt summarize with abstractiveness $a_i$}: '' is used to control abstractiveness, where $a_i \in [0, 100]$ is the abstractiveness. In this paper, we measure abstractiveness $a_i$ using the extractive oracle ROUGE scores \cite{ChinYewLin2004ROUGEAP} between a summary $Y_i$ and its document $X_i$ \cite{nallapati2017summarunner}. The extractive oracle (i.e., a couple of sentences in $X_i$) is obtained by greedily selecting the sentence in $X_i$ that can maximize the ROUGE score w.r.t. the summary $Y_i$ until there is no improvement in ROUGE. Let $Oracle(X_i, Y_i)$ denote the extractive oracle ROUGE score, we set $a_i = 100 * (1 - Oracle(X_i, Y_i))$. Similar to the length control signal, we use bins of size 5 for $a_i$ to cope with sparsity.

\paragraph{Number of Sentences} Controlling the number of sentences in summaries is also useful for users, since sometimes specifying the number of tokens as in \emph{length control} is too difficult for users. The prompt ``{\tt summarize with $n_i$ sentences: }'' is used for controlling lengths of summaries in number of sentences $n_i$. Note that similar to length $l_i$, gold abstractiveness $a_i$ and number of sentences $n_i$ are given during training and during testing, $a_i$ and $n_i$ are given by users, which can basically be any value appearing in training data. Also note that all control signals above are \emph{numbers} and in the following we introduce the control signals that are \emph{words}.

\paragraph{Keywords} Apart from numbers, keywords control can be used to express more general user intents. We use the prompt ``{\tt keywords ($k_1$, $k_2$, $\dots$, $k_{m_i}$) summarize: }'' to generate summaries related to keywords $k_1$, $k_2$, $\dots$, $k_{m_i}$. For example, in an article describing a basketball game between Orlando Magic and Houston Rockets, ``{\tt keywords (Wood blocks) summarize:}'' can be used to guide the model to produce a summary regarding the block performance of Christian Wood.

Following \newcite{he2020ctrlsum}, we obtain the \emph{gold} keywords for model training as follows. We first obtain the extractive oracle sentences using the greedy selection method in \newcite{nallapati2017summarunner} (similar to the \emph{Abstractiveness} control). Then we extract the longest sub-sequences between the gold summary and the extractive oracle sentences obtained earlier, which are the candidate keywords. The gold keywords are these candidates after filtering (i.e., removing stop words and duplicates). The input keywords during test time can be any words (relating to the original documents).
Note that in theory, keywords that are not related to the original document can also be used as input, but the output summary might be meaningless.

\paragraph{Entities}
Similar to keywords, we can also do \emph{entity-centered} generation using the following prompt:
\begin{center}
    {\tt summarize with entities $ET_1(e_{1,1}, e_{1,2}, \dots)$ $ET_2(e_{2,1}, e_{2,2}, \dots)$ $\dots$ }
\end{center}
where $ET_i$ is an entity type (i.e., Location, Organization, Person, etc) and $e_{i,j}$ is an entity of type $ET_i$. For instance, we can use the prompt
    {\tt summarize with entities ORG(Lufthansa) PERSON(Andreas Lubitz)}
to generate summaries on {\tt Lufthansa} and {\tt Andreas Lubitz}.

We use entities appeared in summaries as \emph{gold} entities for training. These entities are extracted using an off-the-shelf named entity recognition toolkit (i.e., spaCy \cite{spacy2}). During test time, any named entity in the input document can be used.

\subsection{Latent Prompt for Summarization}

\begin{table}[!ht]
\small
\begin{center}
\begin{tabular}{|l|c|}
\hline
\textbf{Models} & CNNDM\\
&ROUGE-1/2/L \\

\hline
\hline

\underline{\textbf{Uncontrollable Models}}&\\
BART
&44.62/21.55/41.36 \\

\hline
\hline

\underline{\textbf{Controllable Models}}&\\

\textsc{Length} 
& 45.71/22.13/42.23 \\

\textsc{Abstractiveness} 
& 44.99/21.89/41.86  \\

\textsc{Number of Sentences} 
& 44.47/21.51/41.19  \\

\textsc{Keywords} 
& \textbf{57.32/33.53/53.76} \\

\textsc{Entities} 
& 56.01/32.42/52.49 \\

\hline

\end{tabular}
\end{center}
\caption{Results of Controllable Models with oracle control signals on CNNDM. All models are initialized from BART.}
\label{tab:unctrl_vs_ctrl}
\end{table}

We can achieve \emph{length}, \emph{abstractiveness}, \emph{number of sentences}, \emph{keywords} and \emph{entities} control using prompts and the training methods described in Section \ref{sec:ctrlsum}. However, (gold) control signals may not be available during test time and then the controllable model falls back to an uncontrollable model. We observed in experiments that with gold control signals controllable models achieve much better summarization results compared to their uncontrollable counterparts (see Table \ref{tab:unctrl_vs_ctrl}). This is not surprising since more information of the gold summaries (e.g., length, entities) are given. Therefore, we propose to learn from a controllable model during training phase, so that the control information can be injected into the latent space of the uncontrollable model. 

\paragraph{Latent Prompt} Inspired by GPT-3 \cite{brown2020language}, we use different prompts to distinguish controllable and uncontrollable models (while they share the same model parameters). We use the prompt $Z=\text{\tt summarize:}$ for uncontrollable model. We call $Z$ the \emph{latent prompt}, because it does not contain any control signals but it learns the control signals in the latent space from the controllable model during training.

Let $X=\left(x_1, x_2, \ldots, x_{|X|}\right)$ denote a document and its gold summary $Y=\left(y_1, y_2, \ldots, y_{|Y|}\right)$. Let $P$ denote the prompt for a controllable summarization model (see more details of $P=(q_1, q_2, \dots, q_{|P|})$ in Section \ref{sec:ctrlsum}) and $Z=(z_1, z_2, \dots, z_{|Z|})$ the latent prompt for a uncontrollable model. As shown in Figure \ref{fig:arch}, after prepending these two prompts above to the original document $X$, we obtain transformed documents $X_{c} = P \bigoplus X$ and $X_l = Z \bigoplus X$, where $\bigoplus$ means concatenation.

Then we use a Transformer model parameterized by $\theta$ to take transformed documents $X_c$ and $X_l$ as input and probability distributions for generation the next token $p(y_t | y_{<t}, X_c; \theta)$ and $p(y_t | y_{<t}, X_l; \theta)$ are obtained (also see Equation \ref{eq:model_prob}). We intend to let the uncontrollable model to learn from the distribution of the controllable model using a contrastive learning loss (for bravery we omit $\theta$ in probablities):
\begin{equation}
    \setlength{\abovedisplayskip}{12pt}
    \label{eq:kl_c_teacher}
    \begin{multlined}
        \mathcal{L}^{\text{CL}}(X_c, X_l) \\
        = \sum_{t=1}^{|Y|} D_{\text{KL}}( \text{\tt SG}( p(y_t | y_{<t}, X_c) )|| p(y_t | y_{<t}, X_l))
    \end{multlined}
\end{equation}
where $D_{\text{KL}}$ stands for the Kullback–Leibler divergence and {\tt SG} is the stop gradient operation. We use the stop gradient operation to prevent our model from trivial solutions \cite{xie2020unsupervised}.

Also note that the controllable model may overfit to its prompt $P$ since it may contains too much information of the gold summary $Y$. Therefore, we propose another loss to let the controllable learn from the uncontrollable model as follows:
\begin{equation}
    \label{eq:kl_l_teacher}
    \begin{multlined}
        \mathcal{L}^{\text{CL}}(X_l, X_c) \\
        = \sum_{t=1}^{|Y|} D_{\text{KL}}( \text{\tt SG}( p(y_t | y_{<t}, X_l) )|| p(y_t | y_{<t}, X_c))
    \end{multlined}
\end{equation}

Apart from the contrastive based losses described above, the controllable and uncontrollable models also need to learn from the gold summaries using the negative log-likelihood (NLL) losses (also see Equation \ref{eq:nll_loss}).
\begin{equation}
    \label{eq:nll_loss_c}
    \mathcal{L}^{\text{NLL}}_c(\theta)=-\frac{1}{|Y|} \log p(Y \mid X_c ; \theta)
\end{equation}

\begin{equation}
    \label{eq:nll_loss_l}
    \mathcal{L}^{\text{NLL}}_l(\theta)=-\frac{1}{|Y|} \log p(Y \mid X_l ; \theta)
\end{equation}

Then the final loss the weighted sum of the four losses as follows:
\begin{equation}
    \label{eq:final_loss}
    \begin{multlined}
        \mathcal{L} = \mathcal{L}^{\text{NLL}}_c(\theta) + \mathcal{L}^{\text{NLL}}_l(\theta) \\ 
        + \lambda_1 \mathcal{L}^{\text{CL}}(X_c, X_l) + \lambda_2 \mathcal{L}^{\text{CL}}(X_l, X_c) 
    \end{multlined}
\end{equation}
where $\lambda_1$ and $\lambda_2$ are regularization weights.

\label{sec:multictrl}
\paragraph{Training with Multiple Control Signals} The above process illustrates the training of {\sc Lotus} under a single control signal with control prompt $P$ and latent prompt $Z$. {\sc Lotus} can also be trained with multiple control signals (e.g., \emph{length} and \emph{keywords}). Supposing we have $N$ control signals and their control prompts are $P^{(1)}, P^{(2)}, \dots, P^{(N)}$. During training time, we compute training loss for all prompts. Therefore, the loss for multiple control signals becomes:
\begin{equation}
    \label{eq:multictrl}
    \small
    \begin{multlined}
        \mathcal{L} = \sum_{k=1}^N \mathcal{L}^{\text{NLL}}_{c_k}(\theta) + \mathcal{L}^{\text{NLL}}_l(\theta) \\ 
        + \lambda_1 \sum_{k=1}^N \mathcal{L}^{\text{CL}}(X_{c_k}, X_l) + \lambda_2 \sum_{k=1}^N \mathcal{L}^{\text{CL}}(X_l, X_{c_k}) 
    \end{multlined}
\end{equation}

\section{Experiments}

\subsection{Datasets}

Our experiments are carried out on four popular  summarization datasets i.e., CNN/DailyMail (CNNDM; \citealt{nallapati2016abstractive}), XSum \citep{narayan2018don}, Gigaword \citep{napoles2012annotated} and New York Times (NYT; \citealt{sandhaus2008new}). The remainder of this section introduces the four datasets, and their statistics about control signals are available in Appendix \ref{appendix:ctrl_signals_stat}.

\paragraph{CNNDM} CNNDM contains online news articles paired with reference summaries (associated highlights) from the CNN and DailyMail websites. Following the pre-processing steps introduced in \newcite{see-etal-2017-get},  287,227 document-summary pairs for training, 13,368 for validation and 11,490 for test are obtained.

\paragraph{XSum} XSum aims to create a short (one-sentence) professionally written summary for a given article \cite{narayan2018don}. The articles are collected from BBC online and the summaries are extremely abstractive. The numbers of training, validation and test examples are 204,045, 11,332 and 11,334.

\paragraph{Gigaword} Gigaword is a large corpus of formally edited and annotated English news documents. Given a document, the abstractive headline is used as the summary of its first sentence. We follow the same pre-processing steps in \newcite{rush2015neural} and we obtain 3,803,957 sentence-pairs for training, 189,651 for validation and for 1,951 test.

\paragraph{NYT} The NYT dataset consists of articles published by the New York Times, with summaries written by library scientists. We follow the pre-processing procedures in \newcite{durrett-etal-2016-learning} and \newcite{liu-lapata-2019-text}. There are 38,264 articles for training; 4,002 articles for validation; and 3,421 articles for test in the resulting dataset.

\subsection{Implementation Details}

We adopt BART Large \cite{MichaelLewis2019BARTDS} as our base model. All models are optimized with Adam \cite{kingma2014adam} with $\beta_1=0.9,\beta_2=0.999$. Weight decay is set to 0.01. Learning rates are selected based the validation sets. We use warmup steps 500 across all datasets and \textsc{Lotus} is trained for 20,000/20,000/10,000/6,000 total steps for CNNDM, XSum, Gigaword and NYT, respectively (5 to 10 epochs).

The contrastive loss weights of \textsc{Lotus} (see Equation \ref{eq:final_loss} and \ref{eq:multictrl}) are tuned on validation sets by setting both $\lambda_1$ and $\lambda_2$ to 1 (i.e., $\lambda_1=\lambda_2=1$) or one of them is 1 and the other is 0 (i.e., $\lambda_1=1$ $\lambda_2=0$ or $\lambda_1=0$ $\lambda_2=1$). We use {\tt spaCy} \cite{spacy2} for sentence segmentation and named entity recognition to obtain oracle number of sentences and entity control signals. Oracle keywords control signals are obtained using the greedy keywords extraction implementation in \newcite{he2020ctrlsum}\footnote{https://github.com/salesforce/ctrl-sum}.

\subsection{Results of Uncontrolled Generation}

\begin{table}[!ht]
\small
\begin{center}
\begin{tabular}{|l|c|}
\hline
\textbf{Models} & CNNDM\\
&ROUGE-1/2/L \\

\hline
\hline

\underline{\textbf{Baselines}}&\\

\textsc{BART}
&44.62/21.55/41.36 \\

\hline
\hline

\underline{\textbf{\textsc{Lotus}}}&\\

\textsc{Lotus-Kws} 
&\textbf{45.22/22.09/41.89} \\

\textsc{Lotus-Ent} 
& 45.23/22.02/41.93 \\

\textsc{Lotus-Abs} 
& 45.17/21.97/41.89 \\

\textsc{Lotus-\# Sent} 
& 45.16/21.97/41.86  \\

\textsc{Lotus-Len} 
& 45.15/21.93/41.83 \\

\hline

\end{tabular}
\end{center}
\caption{Uncontrol results of \textsc{Lotus} with single control signal on CNNDM}
\label{tab:single}
\end{table}

\begin{table*}[!ht]
\small
\begin{center}
\begin{tabular}{|l|c|c|c|c|}
\hline
\textbf{Model/Dataset} &CNNDM &XSum &Gigaword &NYT\\
&ROUGE-1/2/L &ROUGE-1/2/L &ROUGE-1/2/L &ROUGE-1/2/L \\

\hline
\hline

\textsc{BERTSUM} \citep{liu2020noisy}
& 42.13/19.60/39.18
& 38.81/16.50/31.27
& --
& 49.02/31.02/45.55
\\

\textsc{T5} \citep{raffel2020exploring}
& 43.52/21.55/40.69
& --
& --
& --
\\

\textsc{PEGASUS} \cite{zhang2020pegasus}
& 44.17/21.47/41.11
& 47.21/24.56/39.25
& --
& --
\\

\textsc{BART} \citep{MichaelLewis2019BARTDS}
& 44.16/21.28/40.90 
& 45.14/22.27/37.25 
& 39.29/20.09/35.65
& --
\\

\textsc{BART} (ours)
& 44.62/21.55/41.36 
& 45.47/22.32/37.01 
& 39.12/20.06/36.37 
& 55.56/36.58/51.17\\

\textsc{CTRLsum} \citep{he2020ctrlsum} 
& 45.65/22.35/42.50 
& -- 
& -- 
& -- \\

\hline
\hline

\underline{\textbf{\textsc{Lotus}}}&&&&\\

\textsc{Lotus-Kws} 
&\textbf{45.31/22.10/41.97} 
&\textbf{45.90/22.66/37.35} 
&39.43/20.40/36.66 
&\textbf{57.73/38.24/53.15}\\

\textsc{Lotus-Len+Abs} 
&45.19/22.02/41.86 
&45.84/22.56/37.28 
&\textbf{39.48/20.52/36.77}
&57.67/38.02/53.10 \\

\textsc{Lotus-All}
& 45.21/22.01/41.90
& 45.85/22.52/37.31
& 39.33/20.28/36.46 
& 57.70/38.06/52.92 \\

\hline

\end{tabular}
\end{center}
\caption{Uncontrol results on four datasets. \textsc{BART} (ours) are our implementation of BART. F1 based ROUGE are reported except for NYT, where limited-length recall based ROUGE is used following \citet{durrett-etal-2016-learning}.}
\label{tab:main}
\end{table*}

In this section, we use ROUGE \cite{ChinYewLin2004ROUGEAP} scores (computed with the perl ROUGE-1.5.5.pl script) to assess the qualities of models in comparison. On CNNDM, XSum and Gigaword, full-length F1 ROUGE 1/2/L are employed, while on NYT, we report limited-length recall based ROUGE 1/2/L following \newcite{durrett-etal-2016-learning}, where model outputs are truncated by their gold summary lengths.

As described in Section \ref{sec:lotus_model}, control signals are not always available during test time, we present results of uncontrolled mode in the following. We first carry out experiments on CNNDM to answer \emph{whether the proposed control signals can help uncontrollable generation?} (see Section \ref{sec:ctrlsum} for details of different control signals).
We compare our model {\sc Lotus} with BART, since {\sc Lotus} is also based on BART, but trained with a different objective.
As shown in Table \ref{tab:single}, all control signals (i.e., \emph{keywords}, \emph{entities}, \emph{abstractiveness}, \emph{number of sentences} and \emph{length}) can help improve uncontrolled results upon BART. It indicates that summarization model with latent prompt can indeed learn from model with control signals, which generates summaries of better quality when control signals are given. Note that {\sc Lotus-Len-Abs} use the prompt ``{\tt summarize with length $l_i$ and abstractivenss $a_i$:}''. Since \mbox{\sc Lotus-Kws} preforms best among non-numerical prompt based models and {\sc Lotus-Len-Abs} performs best among numerical prompt based model. We therefore report results of these two settings on other datasets. Additionally, we also report results of {\sc Lotus-All} to demonstrate that {\sc Lotus} is capable of learning from multiple control signals (i.e., Length, Abstractiveness, Number of Sentences, Keywords, Entities) during training (see the objective in Equation \ref{eq:multictrl}). 

Our main uncontrolled results are summarized in Table \ref{tab:main}. We compare our method against pretrained sequence to sequence Transformer models or pretrained encoder models tailed for text generation, which are T5 \cite{raffel2020exploring}, PEGASUS \cite{zhang2020pegasus}, BART \cite{MichaelLewis2019BARTDS} and BERTSUM \cite{liu2020noisy}. We also compare our model with the uncontrollable results of CTRLsum \cite{he2020ctrlsum}, which firstly extract keywords from a document using a BERT-based \cite{devlin-etal-2019-bert} tagging model and then feed these obtained keywords to a generation model to generate its final summary. By learning from  keywords control signals, \textsc{Lotus-Kws} performs best across three different datasets (i.e., CNNDM, XSum and NYT), which is not surprising since keywords are more expressive than numerical control signals. We also observe in Table \ref{tab:unctrl_vs_ctrl} that with golden keywords, \textsc{Lotus-Kws} obtained much higher ROUGE scores than its numerical control signal based counterparts. Perhaps not surprisingly, \textsc{Lotus-Len+Abs} performs better than \textsc{Lotus-Kws} on Gigaword since source articles of Gigaword are extremely short and cannot contain enough keywords. \textsc{Lotus-All} attemps to model all control signals in a single model, which may ``waste'' too much of its modeling power on controllable generation. As a result, it also performs worse than \textsc{Lotus-Kws}. \textsc{Lotus-Kws} outperforms all models in comparison except for CTRLsum on CNNDM, where our model is slightly worse. Note that CTRLsum is a two stage model, which includes a BERT-based keywords extraction model in additional to its BART based generation model. It is slower than our model and contains more trainable parameters (730M vs 400M). Besides, the keywords model in CTRLsum also leverage thresholds on sentence level and word level, which increases the complexity of implementation and tuning.

\subsection{Results of Controllable Generation}

When control signals are available, {\sc Lotus} can generate text relating to the control signals using prompts we defined in Section \ref{sec:ctrlsum}. we first present results on non-numerical control signals (\emph{keywords} and \emph{entities}) and then results on numerical signals (\emph{length}, \emph{abstractiveness} and \emph{number of sentences}). All results are on the CNNDM test set.

\paragraph{Keywords and Entity Control}
We use the recall of input control keywords or entities to evaluate control abilities of models in comparison. The input control keywords and entities are obtained as follows. We can obtain all oracle keywords and entities from an article (see Section \ref{sec:ctrlsum}). During test time, however, it is not possible for users to provide all keywords or all entities. Therefore, we randomly sample some keywords and entities to simulate user intents\footnote{We first sample the number of keywords $n$ ($n \in [1, 5]$) and then $n$ items from the oracle keywords are sampled.}. Apart from the control capability, we also care about the quality of generated summaries and we therefore also reported the ROUGE-2 score of each model. We compared two of {\sc Lotus} variants ({\sc Lotus-All} and {\sc Lotus-Kws}/{\sc Lotus-Ent}) against CTRLsum\footnote{Available at https://github.com/salesforce/ctrl-sum} \cite{he2020ctrlsum} and CTRLcode \cite{AngelaFan2017ControllableAS}, which prepends control tokens to the beginning of its input document. The original CTRLcode is implemented based on CNNs and for fair comparison we re-implemented it using BART (the same backbone as {\sc Lotus} and CTRLsum).

\begin{table}[!ht]
\centering
\begin{tabular}{|l|c c|}
\hline
Model & Recall & ROUGE-2 \\
\hline
CTRLsum & 0.38 & 23.12 \\
CTRLcode & {\bf 0.51} & 25.09 \\
\hline
\hline
\textsc{Lotus-Kws} & 0.48 & {\bf 25.98} \\
\textsc{Lotus-All} & 0.44 & 25.16 \\
\hline
\end{tabular}

\caption{Keywords control results on CNNDM test set.}
\label{tab:kws_ctrl}
\end{table}

Results of keywords control are reported in Table \ref{tab:kws_ctrl}. {\sc Lotus-Kws} performs much better than CTRLsum and slightly worse than CTRLcode in control ability (see the Recall column). However, {\sc Lotus-Kws} generates summaries with the highest quality (see the ROUGE-2 column) among all models, which indicates our model makes a good trade-off between control ability and summary quality. Perhaps not surprisingly, {\sc Lotus-All}, which accepts all sorts of control signals, does not perform well compared to models trained specifically for keywords control. Results of entity control are reported in Table \ref{tab:ent_ctrl}, {\sc Lotus-Ent} even performs best in control ability and outperforms CTRLsum and CTRLcode in summary quality. {\sc Lotus-All} performs best in summary quality but still worse than {\sc Lotus-Ent} and CTRLcode in control ability. Also note that different from all controllable models in comparison, {\sc Lotus} models can also summarize documents when control signals are not available with good results (see Table \ref{tab:main}). 

\begin{table}[!ht]
\centering
\begin{tabular}{|l|c c|}
\hline
Model & Recall & ROUGE-2 \\
\hline
CTRLsum & 0.36 & 21.97 \\
CTRLcode & 0.60 & 22.22 \\
\hline
\hline
\textsc{Lotus-Ent} & {\bf 0.62} & 22.88 \\
\textsc{Lotus-All} & 0.55 & {\bf 24.42} \\
\hline
\end{tabular}

\caption{Entity control results on CNNDM test set.}
\label{tab:ent_ctrl}
\end{table}

\paragraph{Length, Abs. and \# Sent. Control}

\begin{figure}[!ht]
    \centering
    \includegraphics[width=0.37\textwidth]{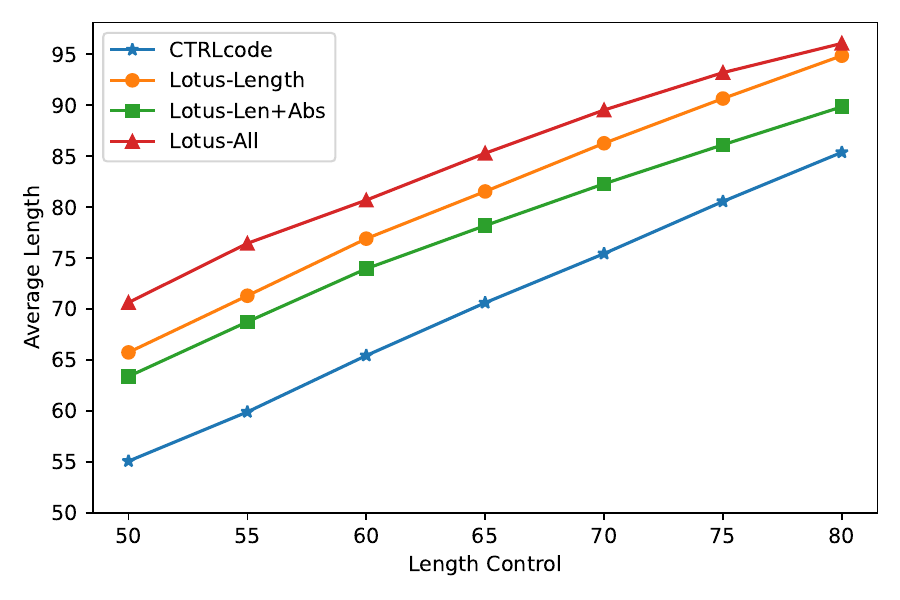}
    \caption{
      Typical length control on CNNDM test set.
    }
    \label{fig:len_arbitrary_ctrl}
\end{figure}

For numerical control signals, we assess control abilities of our models using mean of absolute deviation (MAD; lower the better) between system output and the reference following \newcite{he2020ctrlsum}. A control signal actually stands for one attribute of the model output (e.g., length 30 means the length of the model output should be 30 words). Let $A^{(u)}_i$ denote one user defined control signal (attribute of the desired model output) and $A^{(s)}_i$ the actual attribute of its model output. Then MAD is the mean of all differences between the desired attributes and the actual attributes $\frac{1}{N} \sum_{i=1}^N |A^{(u)}_i - A^{(s)}_i|$, where $N$ is the number of examples for test. As before, we also report ROUGE-2 to evaluate the generation quality. For each test document and each control signal, we use two types of user inputs: the oracle control signal and typical numerical values. For example, in \emph{length} control, we adopt the length of the gold summary and length 50 to 80 with interval 5 (see Figure \ref{fig:len_arbitrary_ctrl}). We compare our model variants with CTRLcode.

\begin{table}[!ht]
\centering
\small
\begin{tabular}{|l|cc|cc|}
\hline
\multirow{2}{*}{Model} & \multicolumn{2}{c|}{Oracle} & \multicolumn{2}{c|}{Typical} \\ 
& MAD & R2 & MAD & R2 \\ \hline
CTRLcode & {\bf 7.45} & 22.13 &  {\bf 5.35} & 21.37 \\
\textsc{Lotus-Len} & 19.09 & 22.34 & 17.23 & 21.65 \\
\textsc{Lotus-Len+Abs} & 12.90 & {\bf 22.36} & 14.60 & {\bf 21.70} \\
\textsc{Lotus-All} & 17.10 & 22.16 & 20.80 & 21.64 \\
\hline
\end{tabular}
\caption{Length control results on CNNDM test set. For MAD, it is lower the better. R2 stands for ROUGE-2.
}
\label{tab:len_ctrl}
\end{table}

The results for length control are in Table \ref{tab:len_ctrl}. Among all {\sc Lotus} models, {\sc Lotus-Len+Abs} performs best (even better than {\sc Lotus-Len}). Perhaps because training examples of different abstractiveness are difficult to be learned without differentiating them, while in the training of {\sc Lotus-Len+Abs}, this problem is alleviated by given the abstractiveness. Although the control ability (see the MAD column) of {\sc Lotus-Len+Abs} is slightly worse than CTRLcode, its output quality is higher. We also show in Figure \ref{fig:len_arbitrary_ctrl} that all models can produce longer output when using a larger length control signal, which demonstrates reasonably good control ability. Abstractiveness and number of sentences control signals follow similar trends and we put these analyses in Appendix \ref{appendix:ctrl_analysis}.

\begin{table*}[!ht]
    \begin{tabular}{|p{\linewidth}|}
    \hline
    \textbf{\textit{Source article}}: (CNN)For the first time in eight years, a TV legend returned to doing what he does best. Contestants told to "come on down!" on the April 1 edition of "The Price Is Right" encountered not host Drew Carey but another familiar face in charge of the proceedings. Instead, there was Bob Barker, who hosted the TV game show for 35 years before stepping down in 2007. Looking spry at 91, Barker handled the first price-guessing game of the show, the classic "Lucky Seven," before turning hosting duties over to Carey, who finished up. Despite being away from the show for most of the past eight years, Barker didn't seem to miss a beat. \\
    \textbf{\textit{Ground-truth summary: }}Bob Barker returned to host "The Price Is Right" on Wednesday. Barker, 91, had retired as host in 2007. \\
    \hline\hline
    \textbf{--- Keywords (\textsc{Lotus-Kws}) ---} \\
    \textbf{Control with user-specified keywords: Bob Barker, return, The Price Is Right} \\
    \textit{{\bf \color{blue}Bob Barker} {\bf \color{blue}returns} to "{\bf \color{blue}The Price Is Right}" for the first time in eight years. The 91-year-old {\bf \color{blue}returned} to the show on April 1. He was on the show for the first time since 2007.} \\
    \hline
   
    \textbf{--- Entities (\textsc{Lotus-Ent}) ---} \\
    \textbf{Control with user-specified entities: Bob Barker, The Price Is Right, 2007} \\
    \textit{{\bf \color{blue}Bob Barker} returns to "{\bf \color{blue}The Price Is Right}" The TV legend has been away from the show since {\bf \color{blue}2007}. He hosted the game show for most of his career before stepping down in {\bf \color{blue}2007}. He was back in charge of the game show's opening game.} \\
    \hline
    
    \textbf{--- Length (\textsc{Lotus-Len}) ---} \\
    \textbf{Uncontrolled mode} \\
    \textit{ Bob Barker returned to "The Price Is Right" for the first time in eight years on April 1. The 91-year-old TV legend hosted the show for 35 years before stepping down in 2007.} \\

    \textbf{Control with length 10} \\ 
    \textit{Bob Barker returned to "The Price Is Right" for the first time in eight years.} \\

    \textbf{Control with length 25} \\ 
    \textit{Bob Barker returned to "The Price Is Right" for the first time in eight years. The 91-year-old TV legend hosted the game show for 35 years.} \\

    \textbf{Control with length 40} \\
    \textit{Bob Barker returned to "The Price Is Right" for the first time in eight years on April 1. The 91-year-old TV legend hosted the game show for 35 years before stepping down in 2007. Drew Carey finished up the show.} \\ 
    \hline
    \end{tabular}
    \caption{Case study of \textsc{Lotus} generated summaries with keywords, entities, and length control.}
    \label{tab:case}
\end{table*}

\subsection{Case Study}
\label{sec:case}

In this section, we present an example in Table \ref{tab:case} to show \textsc{Lotus} generated summaries under keywords, entities, and length control signals. The behaviors of abstractiveness and number of sentences control are similar to that of length control and we put them in Appendix \ref{appendix:abs_sent_num_eg}.

In Table \ref{tab:case}, a sampled article with its ground-truth summary from CNNDM test set is shown in the first block. Both \textsc{Lotus-Kws} (the second block) and \textsc{Lotus-Ent} (the third block) can generate high-quality summaries with user-specified keywords/entities (highlighted in blue and bold). For \textsc{Lotus-Len} (the forth block), in both controlled and uncontrolled mode, high quality summaries are generated (also see the ground-truth summary for comparison). \textsc{Lotus-Len} can also output longer summaries when increasing the length control signal (i.e., 10, 25 and 40).

\section{Conclusions}

In this paper, we first identify the problem that control signals can help to improve the summarization quality, but they are usually unavailable during inference time. Therefore, we propose a Latent Prompt Tuning framework \mbox{\textsc{Lotus}} (i.e., a \textit{single} model with both ``controlled'' and ``uncontrolled'' modes) to solve this problem. Experiments conducted on four datasets show \mbox{\textsc{Lotus}} as a single model can effectively improve the generation quality upon strong (uncontrollable) summarization models. Our analysis further demonstrates \mbox{\textsc{Lotus}} also achieves a good trade-off between control ability and generation quality in controlled mode. In the future, we would like to extend our method to other text generation task such as machine translation, data-to-text generation and large generative language modeling.

\bibliography{anthology,custom}
\bibliographystyle{acl_natbib}

\appendix

\section{Statistics of Control Signals}
\label{appendix:ctrl_signals_stat}

For the control signals introduced in Section \ref{sec:ctrlsum} (i.e., keywords, entities, length, abstractiveness and number of sentences), their statistics on CNNDM, XSum, Gigaword and NYT are reported in Table \ref{tab:statistics}.

\begin{table}[!ht]
\small
\centering
\tabcolsep=0.15cm
\begin{tabular}{|l|llll|}
\hline
\textbf{Control signals} & CNNDM & XSum & Gigaword & NYT \\
\hline
\# of keywords & 11.22 & 2.99 & 2.35 & 16.27 \\
\# of entities & 6.28 & 2.71 & 1.21 & 5.71 \\
Length & 60.46 & 24.15 & 9.19 & 92.74 \\
Abstractiveness & 61.5 & 78.15 & 77.46 & 56.59 \\
\# of sentences & 3.71 & 1.01 & 1.01 & 3.40 \\
\hline
\end{tabular}
\caption{Control signal statistics (average) on CNNDM, XSum, Gigaword and NYT.}
\label{tab:statistics}
\end{table}

\section{Abs. and \# Sent. Control Analysis}
\label{appendix:ctrl_analysis}

In this section, we add more control analysis about abstractiveness and number of sentences control to show more insight about \textsc{Lotus} control ability.

\paragraph{Abstractiveness Control}

\begin{figure}[!ht]
    \centering
    \includegraphics[width=0.37\textwidth]{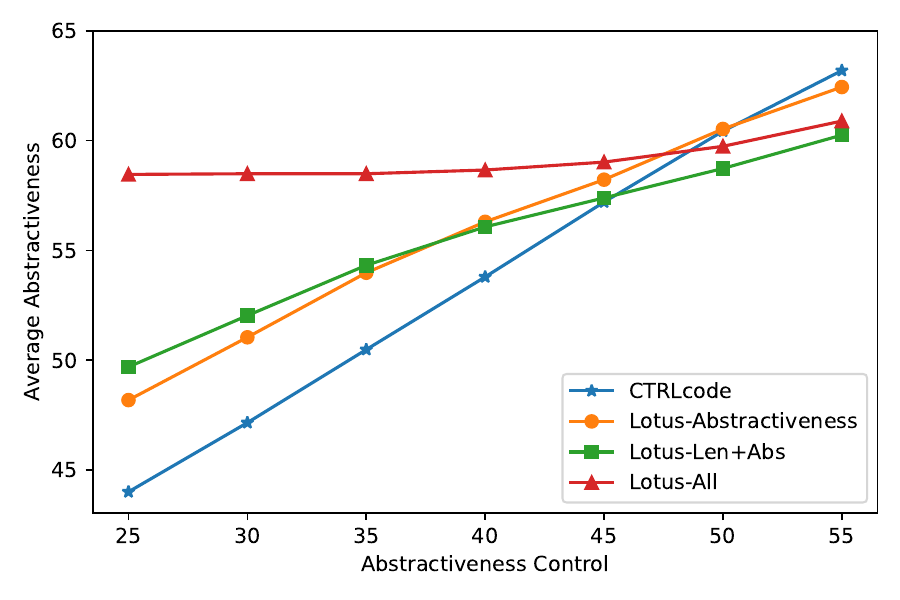}
    \caption{
      Typical Abstractiveness control on CNNDM test set.
    }
    \label{fig:abs_arbitrary_ctrl}
\end{figure}

In \textit{abstractiveness control}, we adopt the abstractiveness of the gold summary as the oracle control and abstractiveness 25 to 55 with interval 5 as the typical control. The results of abstractiveness control are in Table \ref{tab:abs_ctrl}. Among all \textsc{Lotus} models, \textsc{Lotus-Abs} performs best in MAD and ROUGE-2 for oracle control, and best in MAD for typical control. Although its control ability is slightly worse than CTRLcode, its output quality is higher, which is similar to length control. 
\textsc{Lotus-All} performs best in ROUGE-2 for typical control, but its control ability is worse than \textsc{Lotus-Abs}. 
While \textsc{Lotus-Len+Abs} outperforms \textsc{Lotus-Len} on length control, it performs worse than \textsc{Lotus-Abs} on abstractiveness control. Perhaps because we fix the input length control (i.e., $l_i$ in ``{\tt summarize with length $l_i$ and abstractivenss $a_i$:}'') as the average length of gold summaries (oracle length is unknown for abstractiveness control), which may affect \textsc{Lotus-Len+Abs} to generate summaries around the average length instead of gold length and heavily penalize both ROUGE-2 and abstractiveness MAD.
Figure \ref{fig:abs_arbitrary_ctrl} also shows that all models can produce summaries with larger abstractiveness when using a larger abstractiveness control signal.

\begin{table}[!ht]
\centering
\small
\begin{tabular}{|l|cc|cc|}
\hline
\multirow{2}{*}{Model} & \multicolumn{2}{c|}{Oracle} & \multicolumn{2}{c|}{Typical} \\
                       & MAD           & R2          & MAD           & R2           \\ \hline
CTRLcode &{\bf 15.26} &21.89 &{\bf 15.58} &20.98 \\
\textsc{Lotus-Abs} &17.34 &{\bf 22.37} &17.76 &21.71 \\
\textsc{Lotus-Len+Abs} &24.84 &21.51 &17.80 &21.61 \\
\textsc{Lotus-All} &21.78 &21.96 &21.40 &{\bf 21.84} \\
\hline
\end{tabular}
\caption{Abstractiveness control results on CNNDM test set.}
\label{tab:abs_ctrl}
\end{table}

\paragraph{Number of Sentences Control}

In \textit{number of sentences control}, we adopt the number of sentences of the gold summary as the oracle control and number of sentences 1, 2, 3, 4, and 5 as the typical control. {\sc Lotus} performs better than CTRLcode for both oracle and typical control. Specifically, {\sc Lotus-\# Sent} performs best in both control ability (i.e., MAD) and output quality (ROUGE-2) for oracle control. It also performs best in control ability for typical control, while {\sc Lotus-All} performs best in output quality.

\begin{table}[!ht]
\centering
\small
\begin{tabular}{|l|ll|ll|}
\hline
\multirow{2}{*}{Model} & \multicolumn{2}{c|}{Oracle} & \multicolumn{2}{c|}{Typical} \\
& MAD & R2 & MAD & R2 \\ 
\hline
CTRLcode &0.60 &21.51 
&0.72 &19.85 \\
\textsc{Lotus-\# Sent} &{\bf 0.58} &{\bf 22.10} 
&{\bf 0.63} &20.86 \\
\textsc{Lotus-All} &0.60 &21.90 
&0.78 &{\bf 21.22} \\
\hline
\end{tabular}
\caption{Number of sentences control results on CNNDM test set.}
\label{tab:sent_ctrl}
\end{table}

\begin{figure}[!ht]
    \centering
    \includegraphics[width=0.37\textwidth]{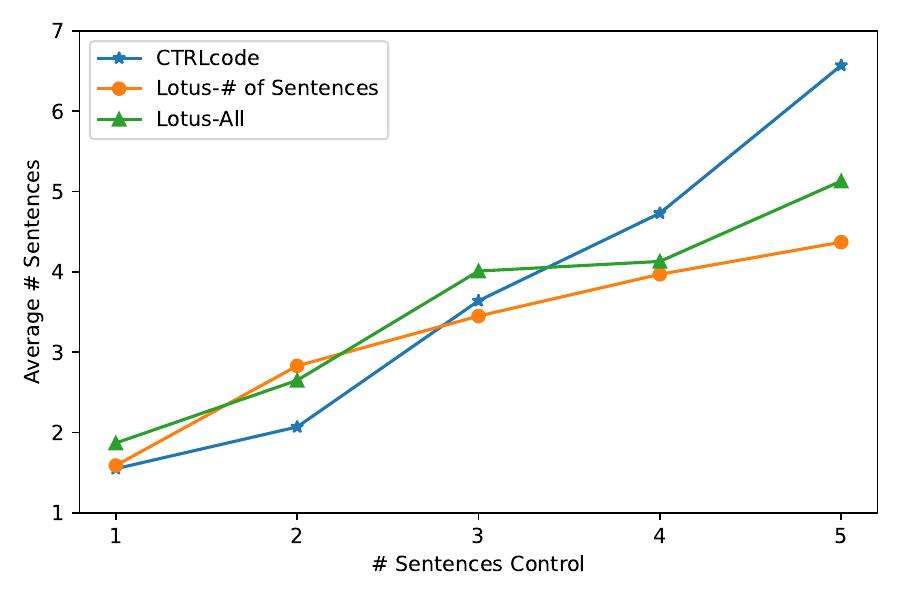}
    \caption{
      Typical number of sentences control on CNNDM test set. 
    }
    \label{fig:sent_arbitrary_ctrl}
\end{figure}

\section{Abs and \# Sent. Case Study}
\label{appendix:abs_sent_num_eg}

Taking the same sampled case in Section \ref{sec:case}, Table \ref{tab:case_abs_sent_num} shows the example outputs of \textsc{Lotus} generated summaries with abstractiveness and number of sentences control. Example outputs of \textsc{Lotus-Abs} and \textsc{Lotus-\# Sent} are similar to \textsc{Lotus-Len}, both controlled and uncontrolled modes can generate high quality summaries. \textsc{Lotus-Abs} can generate more abstractive summaries when increasing the abstractiveness control signals. When the number of sentences control signals increases, \textsc{Lotus} generated summary also has a similar increasing trend and matches the control signals more exactly.

\hfill

\centering
\begin{tabular}[!ht]{|p{0.95\linewidth}|}
    \hline
    \textbf{\textit{Source article}}: (CNN)For the first time in eight years, a TV legend returned to doing what he does best. Contestants told to "come on down!" on the April 1 edition of "The Price Is Right" encountered not host Drew Carey but another familiar face in charge of the proceedings. Instead, there was Bob Barker, who hosted the TV game show for 35 years before stepping down in 2007. Looking spry at 91, Barker handled the first price-guessing game of the show, the classic "Lucky Seven," before turning hosting duties over to Carey, who finished up. Despite being away from the show for most of the past eight years, Barker didn't seem to miss a beat. \\
    \textbf{\textit{Ground-truth summary: }}Bob Barker returned to host "The Price Is Right" on Wednesday. Barker, 91, had retired as host in 2007. \\
    \hline
\end{tabular}

\begin{table}[]
    \begin{tabular}{|p{\linewidth}|}
    \hline
    \textbf{--- Abstractiveness (\textsc{Lotus-Abs}) ---} \\
    \textbf{Uncontrolled} \\
     \textit{Bob Barker returned to "The Price Is Right" for the first time in eight years on April 1. The 91-year-old TV legend hosted the game show for 35 years before stepping down in 2007. Drew Carey finished up hosting the April 1 edition of "The Price Is Right"} \\

    \textbf{Control with abstractiveness 45} \\
    \textit{Bob Barker hosted "The Price Is Right" for the first time in eight years on April 1. He hosted the show for 35 years before stepping down in 2007. The 91-year-old didn't seem to miss a beat, looking spry at 91.} \\
     
     \textbf{Control with abstractiveness 70} \\
    \textit{Bob Barker returned to "The Price Is Right" for the first time in eight years on April 1. The 91-year-old TV legend hosted the game show for 35 years before stepping down in 2007. Drew Carey finished up hosting the show, which Barker had hosted for most of his career.} \\

    \textbf{Control with abstractiveness 95} \\
    \textit{Bob Barker returned to "The Price Is Right" for the first time in eight years. The 91-year-old TV legend has hosted the game show for 35 years. Drew Carey finished up hosting the April 1 edition of the show, which Barker had hosted since 2007.} \\ 
     
    \hline
    \textbf{--- Number of Sentences (\textsc{Lotus-\# Sent}) ---} \\
     \textbf{Uncontrolled} \\
    \textit{ Bob Barker returned to "The Price Is Right" for the first time in eight years. The 91-year-old TV legend hosted the game show for 35 years. Drew Carey finished up the show, which he hosted since 2007.} \\

    \textbf{Control with number of sentences 1} \\
    \textit{Bob Barker returned to "The Price Is Right" for the first time in eight years.} \\
    
    \textbf{Control with number of sentences 2} \\ 
    \textit{Bob Barker returned to "The Price Is Right" for the first time in eight years. The 91-year-old TV legend hosted the game show for 35 years before stepping down in 2007.} \\

    \textbf{Control with number of sentences 3} \\ 
    \textit{Bob Barker returned to "The Price Is Right" for the first time in eight years. The 91-year-old TV legend hosted the game show for 35 years before stepping down in 2007. At 91, Barker didn't seem to miss a beat.} \\
    \hline
    \end{tabular}
    \caption{Case study of \textsc{Lotus} generated summaries with abstractiveness, and number of sentences control.}
    \label{tab:case_abs_sent_num}
\end{table}

\end{document}